\documentclass[letterpaper, 10 pt, conference]{ieeeconf}  

\usepackage{graphicx}
\usepackage{comment}
\usepackage{bm}
\usepackage{amsmath}
\usepackage{amsfonts}
\usepackage{amssymb}
\usepackage{cite}
\usepackage[caption=false,font=footnotesize]{subfig}

\IEEEoverridecommandlockouts                              

\title{\LARGE \bf
Motion Generation Considering Situation with Conditional Generative Adversarial Networks for Throwing Robots*
}

\author{Kyo Kutsuzawa$^{1}$, Hitoshi Kusano$^{2}$, Ayaka Kume$^{2}$, and Shoichiro Yamaguchi$^{2}$ 
\thanks{*This work was supported by Preferred Networks, Inc.}
\thanks{$^{1}$Kyo Kutsuzawa is with JSPS Research Fellow (DC2) and Graduate School of Science and Engineering, Saitama University,
        255 Shimo-Ohkubo, Saitama, Japan
        {\tt\small k.kutsuzawa.430@ms.saitama-u.ac.jp}}%
\thanks{$^{2}$Hitoshi Kusano, Ayaka Kume and Shoichiro Yamaguchi are with Preferred Networks,
        Tokyo, Japan
        {\tt\small \{kusano, kume, guguchi\}@preferred.jp}}%
}

\begin{document}

\maketitle
\thispagestyle{empty}
\pagestyle{empty}


\begin{abstract}
When robots work in a cluttered environment, the constraints for motions change frequently and the required action can change even for the same task.
However, planning complex motions from direct calculation has the risk of resulting in poor performance local optima.
In addition, machine learning approaches often require relearning for novel situations.
In this paper, we propose a method of searching appropriate motions by using conditional Generative Adversarial Networks (cGANs), which can generate motions based on the conditions by mimicking training datasets.
By training cGANs with various motions for a task, its latent space is fulfilled with the valid motions for the task.
The appropriate motions can be found efficiently by searching the latent space of the trained cGANs instead of the motion space, while avoiding poor local optima.
We demonstrate that the proposed method successfully works for an object-throwing task to given target positions in both numerical simulation and real-robot experiments.
The proposed method resulted in three times higher accuracy with 2.5 times faster calculation time than searching the action space directly.
\end{abstract}

\section{Introduction}
In the near future, robots are expected to work in our daily lives, which are often cluttered with objects such as furniture \cite{Bugmann2011}.
They should execute various tasks, such as cooking meals, cleaning rooms, and carrying dishes, while avoiding obstacles.

Many motion planning methods have been proposed to obtain robot motions for such complex tasks.
For example, Model Predictive Control (MPC) \cite{Garcia1989,Findeisen2002,Rao2014} is a common approach for the motion optimization.
This method has been applied to self-driving vehicles\cite{Paden2016}, aerial vehicles\cite{Garimella2015}, and humanoid robots\cite{Audren2014}.
For tasks in which the derivatives of the models are unknown, sampling-based methods are also studied \cite{Williams2016,Williams2018}.
Such optimization methods, however, should search for solutions from all possible motions.
Searching all possible motions carries a risk of bad initial conditions resulting in a poor performance local optimum \cite{Kelman2011,Daehlen2014}.

Machine learning approaches based on deep neural networks have also been studied widely \cite{Pierson2017}.
This approach can be used for complex tasks that are difficult to solve with analytical methods \cite{Lenz2015}.
In addition, once a neural network is trained, the neural network can generate motions with lower computational costs than optimization from scratch \cite{Zhang2016}.
Neural networks can be trained from various sources such as labeled images\cite{Lenz2015a}, human demonstrations\cite{Ho2016,Duan2017,Yang2017}, and other optimization results \cite{Zhang2016}.
Deep neural networks, however, require a large amount of training data.
It is difficult to prepare a large training dataset in which situations and their optimal motions are associated.
Moreover, generalization may not be possible for novel situations that did not appear in the training dataset.
In addition to supervised learning, reinforcement learning has also been applied to learn the association between situations and actions \cite{Peng2017,Gu2017,Zeng2018a}.
However, reinforcement learning is difficult to train and must often be retrained for situations not encountered during training.
To achieve a high generalization performance for a variety of situations, many trials are necessary.
Some studies \cite{pinto2017robust,rajeswaran2016epopt} aim to learn robust policies for condition changes, such as physical parameters, but they cannot adapt to totally new situations.

To summarize the above discussion, it is difficult to plan motions for complex tasks in various situations.
Generic motion optimization methods have a risk of poor local optima, because they search solutions from all the motions including inappropriate conditions.
On the other hand, learning-based methods require relearning for novel situations.

In this paper, we propose a motion planning method based on conditional Generative Adversarial Networks (cGANs)\cite{Mirza2014a}, which generates various motions from given conditions and latent variables by mimicking training datasets.
cGANs do not require designing the characteristics of the latent space by hand, unlike other methods\cite{Cully2015}.
The proposed method aims to divide the motion planning into two phases:
1) training various motions to cGANs regardless of the relationship to their situations and
2) searching appropriate motions for given situations from the latent space of the trained cGANs that is associated only with valid motions.
The proposed method searches for the best motion from only valid motions represented in the latent space, avoiding possible convergence on an invalid motion.
In addition, relearning is not required, because cGANs can generate motions suitable for specific situations.

While cGANs can generate various kinds of motions, in this paper, we look at a task where a robot throws objects to target positions positions.
Throwing enables object movement beyond the robots' reachable spaces rapidly; in living spaces cluttered with objects, there may be places where robots cannot enter or reach.
However, it is more difficult to plan throwing motions than in the normal object placement tasks.
This is because the contact condition between the robot and the object changes discontinuously based on the motions, which makes the dynamics complex.
In this task, there exist various possible throwing motions to reach the target and robots need to choose a motion according to the situation.
We cannot determine object trajectories and robot motions uniquely even by giving target positions.
Therefore, it is not enough to consider end-effector positions and parabolas of the object.
Although many studies tackled the robotic throwing task in the aspects of model-based approach \cite{Tabata2002,Miyashita2009,Sintov2015,Kim2017} and learning-based approach \cite{Kober2012,DaSilva2014}, they were unable deal with novel situations without replanning from scratch or relearning.


\section{Background} \label{sec:basis}
In this section, we introduce \emph{conditional Wasserstein GANs with Gradient Penalty} (cWGANs-GP) and \emph{Covariance Matrix Adaptation Evolution Strategy} (CMA-ES).

\subsection{cGANs}
cGANs are generative models that can generate various samples corresponding to condition inputs, by mimicking training datasets \cite{Mirza2014a}.
cGANs are extension of GANs, which are composed of a generator network and a discriminator network \cite{Goodfellow2014}.
The generator network aims to generate samples imitating a given dataset, while the discriminator network aims to distinguish generated data and actual data in the dataset.
The generator is implemented as a deep neural network that maps a condition input $\bm{c}$ and a random variable $\bm{z}$ sampled from a uniform distribution to data $\hat{\bm{x}}$.
The discriminator is also implemented as a deep neural network which distinguishes the actual data $\bm{x}$ from the dataset and the generated sample $\hat{\bm{x}}$ from the generator.


Wasserstein GANs (WGANs) \cite{Arjovsky2017a} are kinds of GANs that use Wasserstein distance to measure the difference of probability distributions between the training dataset and the generated data.
WGANs are stable to learning the probability distributions thanks to the use of Wasserstein distance instead of Jensen-Shannon divergence, which is used in the generic GANs \cite{Arjovsky2017}.
In addition, WGANs can be improved by applying a penalty term called the gradient penalty during training \cite{Gulrajani2017}.
Such WGANs are called as WGANs-GP.
Parameters of the generator and discriminator of WGANs-GP are optimized by the following loss functions:
\begin{align}
    L_G =& -\mathbb{E}_{p(\bm{z})}[D(G(\bm{z}))], \\
    L_D =& \mathbb{E}_{p(\bm{z})}[D(G(\bm{z}))] - \mathbb{E}_{p(\bm{x})}[D(\bm{x})] \nonumber\\
    &+ \lambda \mathbb{E}_{p(\hat{\bm{x}})}[(\|\nabla_{\hat{\bm{x}}}D(\hat{\bm{x}})\|_2 - 1)^2],
\end{align}
where
\begin{equation}
    \hat{\bm{x}} = \epsilon \bm{x} + (1-\epsilon) G(\bm{z}), \quad \epsilon \sim U[0, 1].
\end{equation}
Here, $G$ and $D$ are the generator and the discriminator, respectively.
$\mathbb{E}_{p(\bullet)}$ denotes expectation under the probability $p(\bullet)$.
$L_\bullet$ is the training loss, $\bm{x}$ is a sample in the training dataset, and $U$ indicates the uniform distribution.
$\lambda$ is a hyperparameter for the gradient penalty.

cGANs and WGANs-GP can be combined.
Such models are called cWGANs-GP.
The simplest implementation of the cWGANs-GP is to input the conditions $\bm{c}$ to both the generators and the discriminators of WGANs-GP.
The loss function is expressed as follows:
\begin{align}
    L_G =& -\mathbb{E}_{p(\bm{c})}\left[\mathbb{E}_{p(\bm{z})}[D(G(\bm{z}, \bm{c}), \bm{c})]\right], \\
    L_D =& \mathbb{E}_{p(\bm{c})}\left[\mathbb{E}_{p(\bm{z})}[D(G(\bm{z}, \bm{c}), \bm{c})]\right] \nonumber\\
    &- \mathbb{E}_{p(\bm{c})}\left[\mathbb{E}_{p(\bm{x|\bm{c}})}[D(\bm{x}, \bm{c})]\right] \nonumber\\
         &+ \lambda \mathbb{E}_{p(\bm{c})}\left[\mathbb{E}_{p(\hat{\bm{x}}|\bm{c})}[(\|\nabla_{\hat{\bm{x}}}D(\hat{\bm{x}}, \bm{c})\|_2 - 1)^2]\right],
\end{align}
where
\begin{equation}
    \hat{\bm{x}} = \epsilon \bm{x} + (1-\epsilon) G(\bm{z}, \bm{c}), \quad \epsilon \sim U[0, 1].
\end{equation}
Here, $\bm{c}$ denotes the condition of the training sample $\bm{x}$.

\subsection{CMA-ES} \label{sec:cma-es}
CMA-ES is a stochastic optimization method for nonlinear, nonconvex functions \cite{Hansen2001,Hansen2016}.

This method can be applied even when the objective functions are multimodal and ill-scaled.
In addition, all hyperparameters have recommended values which are only depend on the number of dimension \cite{Hansen2014}.
Therefore, CMA-ES is expected to work regardless of the geometry of the objective functions in the searching space.
Moreover, because CMA-ES is a gradient-free method, it can be used for even nondifferentiable objective functions.

The search in CMA-ES progresses based on a Gaussian distribution.
Their mean vector and covariance matrix are updated iteratively according to the evaluation values of the samples from the Gaussian distribution.

\section{Method} \label{sec:method}
This section explains the proposed method by using the throwing task as an example.
The proposed method handles the issue by dividing it into two phases:
1) letting a generative model learn various motions and
2) finding a motion optimized for a given situation from the trained generative model.

\subsection{Generative Model of Various Motions}
To train various motions, we use cWGANs-GP.
The data to be generated by the cWGANs-GP, $\bm{x}$, is the motion of the robots for the task.
Motion primitives can be used to represent motions as explained in Section~\ref{sec:setting}.
The condition $\bm{c}$ is the goal of the task.
The design of the conditions depends on the task.
For example, in the placing tasks, target positions to place objects can be candidates of the condition.



The generated samples of cGANs depend on the quality of training datasets.
The proposed method requires diverse and valid motions.
To obtain these motions for the training datasets, we generated random actions and used the actions for training if they satisfy the target task’s condition.




\subsection{Searching in Latent Space}
After the cGANs is trained, valid motions can be found by searching the latent space.
Thanks to the latent variables expressing only valid motions, it is not necessary to filtering out inappropriate motions for the task.
We use CMA-ES to search the latent spaces due to the good properties explained in Section~\ref{sec:cma-es}.
It should be noted that other searching methods can be used instead of CMA-ES.

The search process consists of sampling latent variables $\bm{z}$ and evaluating $\bm{x}=G(\bm{z})$ with an arbitrary objective function according to the situation.
The algorithm is as follows:
\begin{enumerate}
    \item Define an objective function.
    \item Sample latent variables $\bm{z}$ from the latent space according to the parameters of CMA-ES.
    \item Generate motions $\bm{x}=G(\bm{z})$ and evaluate them with the objective function.
    \item Update parameters of CMA-ES based on the values of the objective function.
    \item Repeat the above steps until a suitable solution is found.
\end{enumerate}

During searching latent variables, CMA-ES samples solution candidates from the whole latent space.
However, $\bm{z}$ should be sampled from the range of $-1$ to $1$.
Therefore, we applied $\tanh$ function to the solution candidates to limit the range of $\bm{z}$.



\section{Task Specification} \label{sec:setting}
\begin{figure}
    \centering
    \includegraphics[scale=0.5]{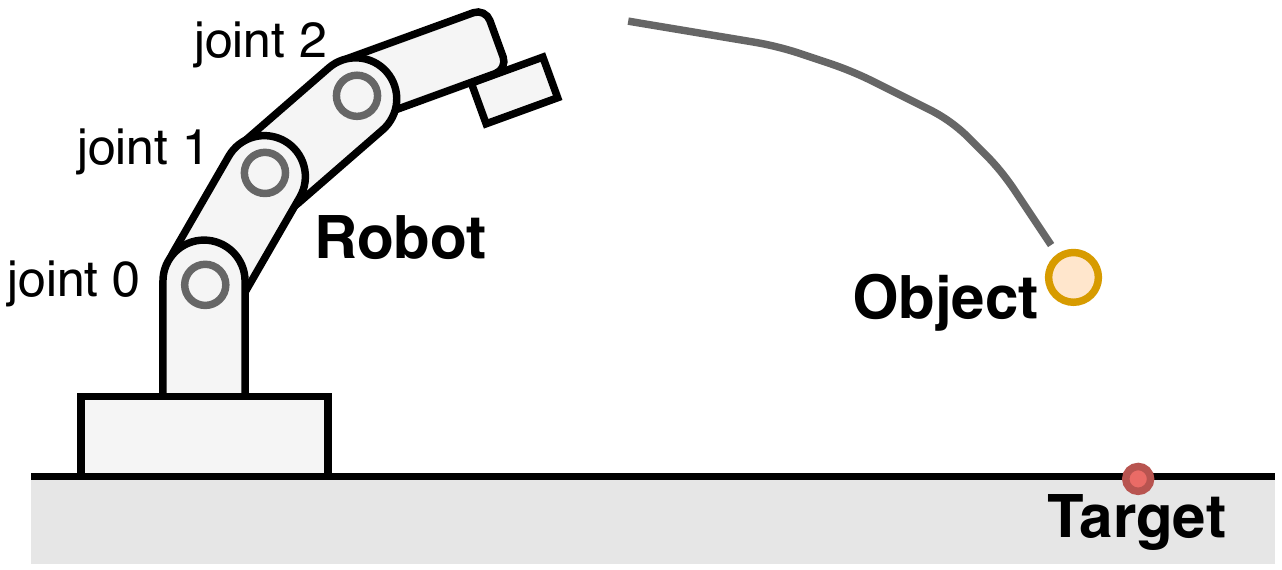}
    \caption{Overview of the task considered in this paper.}
    \label{fig:task overview}
\end{figure}

This paper takes the task of robotic throwing as illustrated in Fig.~\ref{fig:task overview}.

\subsection{Physics Model} \label{sec:physics model}
This paper considers throwing motions by a manipulator with three degrees of freedom (DoF) in planar physics.

The state variables of the robot are defined as follows:
\begin{equation}
    \bm{x}(t) = [\theta_1(t), \theta_2(t), \theta_3(t), \dot{\theta}_1(t), \dot{\theta}_2(t), \dot{\theta}_3(t)]^\top.
\end{equation}
The state variables follow the following state equation:
\begin{equation}
    \bm{x}(t+1) = \bm{A} \bm{x}(t) + \bm{B} \bm{u}(t),
\end{equation}
where
\begin{align}
    \bm{u}(t) &= [\ddot{\theta}_1(t), \ddot{\theta}_2(t), \ddot{\theta}_3(t)]^\top ,\\
    \bm{A} &=
    \begin{bmatrix}
        \bm{I}_3  &\Delta t \bm{I}_3\\
        \bm{O}_3  &\bm{I}_3
    \end{bmatrix},\\
    \bm{B} &=
    \begin{bmatrix}
        \bm{I}_3 \frac{\Delta t^2}{2}\\
        \bm{I}_3
    \end{bmatrix}.
\end{align}
Here, $\Delta t$ denotes the sampling interval.
$\bm{I}_3 \in \mathbb{R}^{3 \times 3}$ and $\bm{O}_3 \in \mathbb{R}^{3 \times 3}$ are the identity matrix and the zero matrix, respectively.

The object is described as a mass point.
Therefore, the motion can be described as follows:
\begin{equation}
    \bm{x}_\mathrm{o}(t) = [p_x(t), p_y(t), \dot{p}_x(t), \dot{p}_y(t)]^\top,
\end{equation}
and
\begin{equation}
    \bm{x}_\mathrm{o}(t+1) = \bm{A}_\mathrm{o} \bm{x}_\mathrm{o}(t) + \bm{B}_\mathrm{o} \bm{u}_\mathrm{o}(t),
\end{equation}
where
\begin{align}
    \bm{u}_\mathrm{o}(t) &= [\ddot{p}_x(t), \ddot{p}_y(t)]^\top ,\\
    \bm{A}_\mathrm{o} &=
    \begin{bmatrix}
        \bm{I}_2  &\Delta t \bm{I}_2\\
        \bm{O}_2  &\bm{I}_2
    \end{bmatrix},\\
    \bm{B}_\mathrm{o} &=
    \begin{bmatrix}
        \bm{I}_2 \frac{\Delta t^2}{2}\\
        \bm{I}_2
    \end{bmatrix}.
\end{align}

The manipulator has a bowl-shaped end-effector at the tip of the arm.
The contact models are expressed as follows:
\begin{align}
    \bm{u}_\mathrm{o} = 
    \begin{cases}
        \bm{a}_\mathrm{e}  &\mathrm{if\ } \bm{n}_\mathrm{e} \cdot (\bm{a}_\mathrm{e}-\bm{g}) \geq 0,\\
        \bm{g}    &\mathrm{if\ } \bm{n}_\mathrm{e} \cdot (\bm{a}_\mathrm{e}-\bm{g}) < 0.\\
    \end{cases}
\end{align}
Here, $\bm{a}_\mathrm{e}$ is the acceleration of the end-effector and $\bm{g}$ is the gravity.
$\bm{n}_\mathrm{e}$ is the normal vector to the end-effector.
Thus, the object is constrained to the end-effector until $\bm{n}_\mathrm{e} \cdot (\bm{a}_\mathrm{e}-\bm{g}) \geq 0$.

\subsection{Representation of Motions}
Although the manipulator can be controlled by specifying $\bm{x}(t)$, it is difficult for neural networks to handle the time series of the state variables directly \cite{Pascanu2013}.
To reduce the burden of the neural networks, we employ the idea of motion primitives, which represent complex motions as the combinations of simple primitive motions.

\begin{figure}
    \centering
    \includegraphics[width=7.0cm]{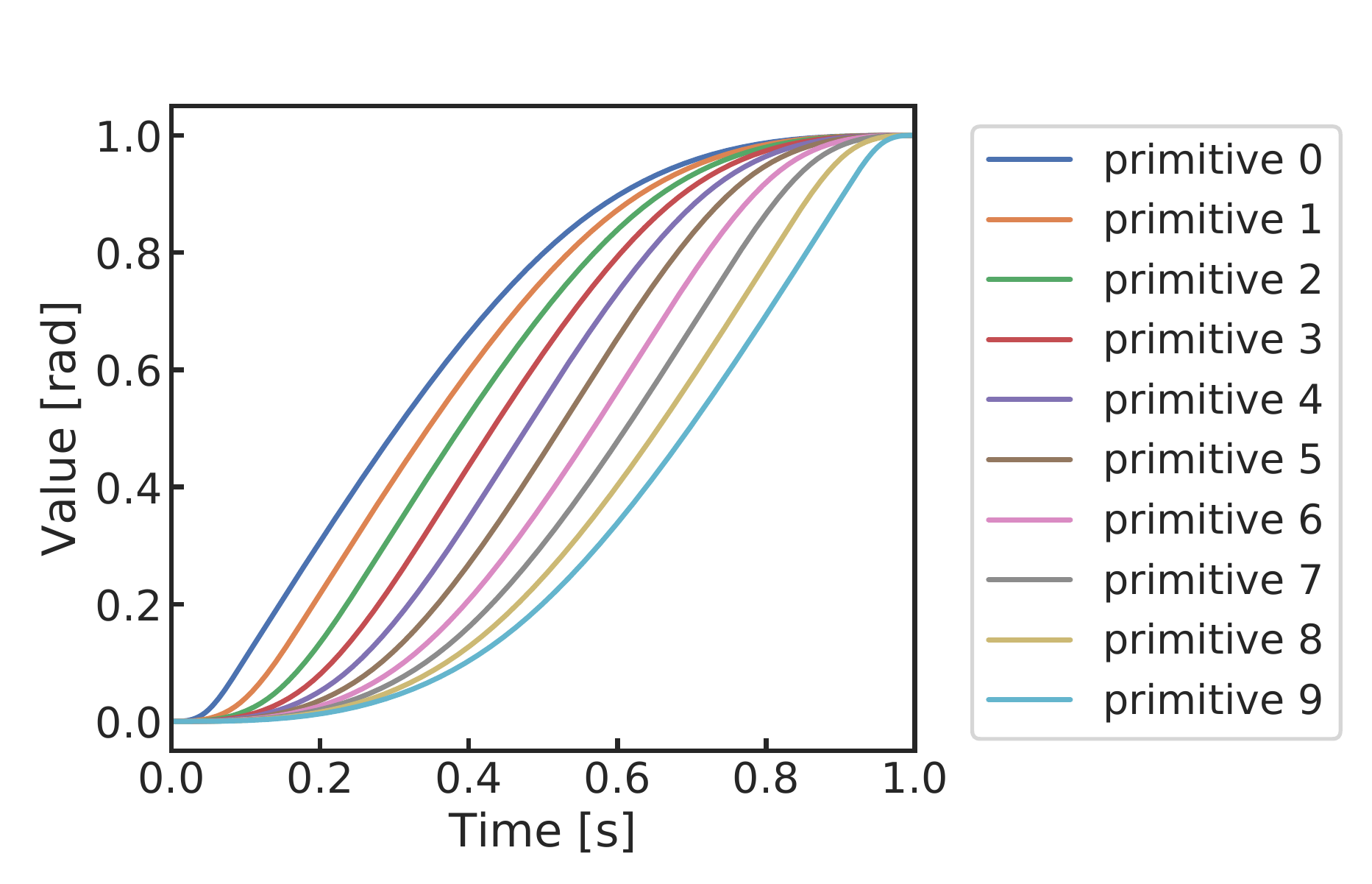}
    \caption{Motion primitives.}
    \label{fig:motion primitives}
\end{figure}

Motions of the manipulator are expressed as linear combinations of motion primitives.
The $i$-th joint angle at time $t$, $\theta_i (t)$, is expressed as follows:
\begin{equation}
    \theta_i (t) = \theta_i^\mathrm{init} + \sum_{j=0}^{J-1} w_{ij} \phi_j(t).
\end{equation}
Here, $\theta_i^\mathrm{init}$ is the initial joint angle and $w_{ij}$ is a weight for the primitives.
$\phi_j$ is an S-curve defined as follows:
\begin{equation}
    \frac{\mathrm{d}^3}{\mathrm{d}t^3} \phi_j(t) =
    \begin{cases}
        \frac{8}{\tau_j^2T}   & 0 \leq t < \frac{\tau_j}{2}\\
        -\frac{8}{\tau_j^2T}  & \frac{\tau_j}{2} \leq t < \tau_j\\
        \frac{8}{\tau_j^2T}   & \tau_j \leq t < \frac{\tau_j+T}{2}\\
        -\frac{8}{\tau_j^2T}  & \frac{\tau_j+T}{2} \leq t < T\\
    \end{cases},
\end{equation}
where,
\begin{equation}
    \tau_j = \frac{j+1}{J+2}T,
\end{equation}
Here, $T$ is the length of the motions and $J$ is the number of primitives.
Fig.~\ref{fig:motion primitives} shows the primitives with $J=10$ and $T=1.0 \mathrm{\ s}$, which are used in the experiments.

\section{Results of Training cGANs} \label{sec:training}
This section explains the details of the dataset, model implementation, and results of the training.

\subsection{Dataset}
A large amount of training data is necessary to train cGANs.
Because it is not easy to obtain large datasets with actual robots, we used a simulator based on the physics model explained in Section~\ref{sec:physics model} to generate throwing motions.

\begin{figure}
    \centering
    \subfloat[target position: 84.8 cm]{
        \includegraphics[width=8cm]{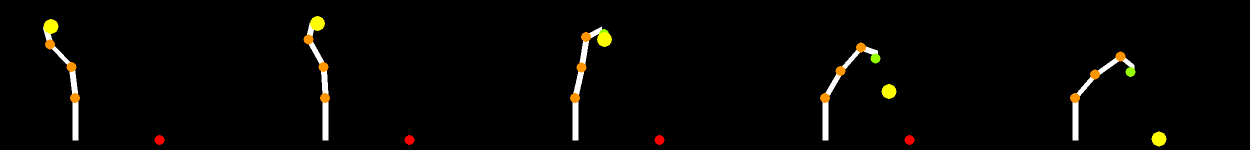}
    }\\
    \subfloat[target position: 128.8 cm]{
        \includegraphics[width=8cm]{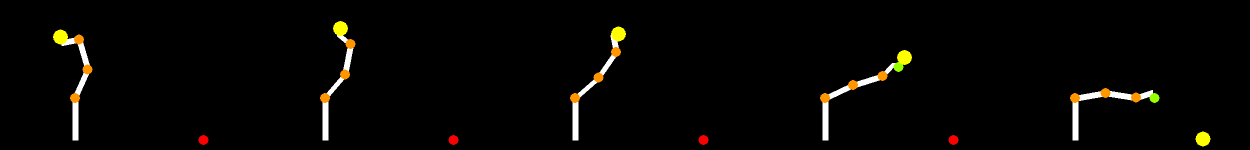}
    }\\
    \subfloat[target position: 99.2 cm]{
        \includegraphics[width=8cm]{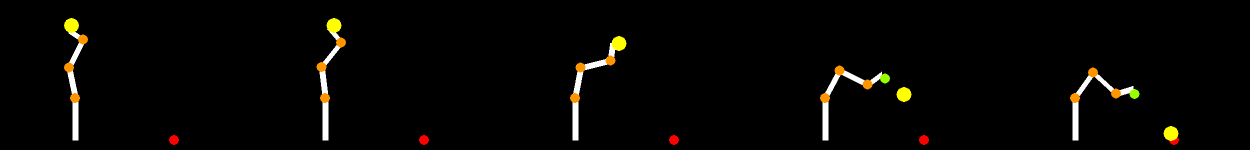}
    }
    \caption{Snapshots of throwing motions in the training data.
        The yellow circles indicate the objects.
        The red circles indicate the landing positions.}
    \label{fig:snapshots of training data}
\end{figure}

The dataset was obtained in a self-supervised manner with the following procedure.
At first, a random initial pose $\bm{x}(0)$ and a random action $\bm{w}$ are generated.
Here, both $\bm{x}(0)$ and $\bm{w}$ were generated from the uniform distribution.
$\bm{x}(0)$ was sampled from the ranges in Table~\ref{tbl:limitations}, while each component of $\bm{w}$ was sampled from $[-2, 2)$.
Then, the action $\bm{w}$ is simulated to obtain the flying distance, $p_\mathrm{g}$.
After that, $\bm{x}(0)$, $\bm{w}$, and $p_\mathrm{g}$ are added to the dataset if they are valid.
The validity is defined as follows:
\begin{itemize}
    \item The flying distance $p_\mathrm{g}$ is in the range 0.8--1.4 m.
        Because the reachable distance of the robot is 0.78 m, this range can be reached only by the throwing motions.
    \item The contact between the end-effector and the object is maintained at $t = 0$.
    \item The motion does not exceed the limitation in Table~\ref{tbl:limitations}.
\end{itemize}
Finally, we reduced the obtained data to level the occurrence of the target positions evenly.
In total, we obtained 282500 throwing motions.
Examples of the training data are shown in Fig.~\ref{fig:snapshots of training data}.

\subsection{Implementation}
\begin{figure}
    \centering
    \subfloat[Generator]{
        \includegraphics[scale=0.35]{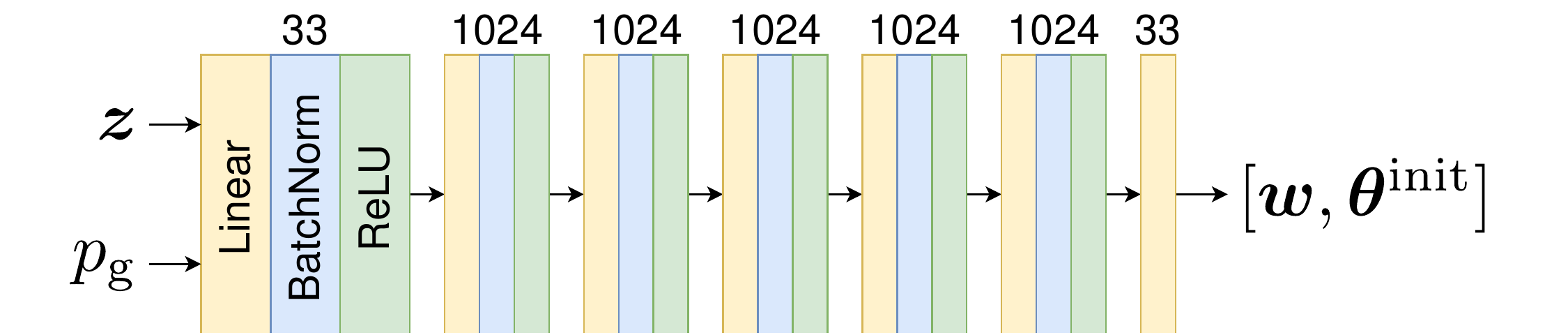}
        \label{fig:generator}}\\
    \subfloat[Discriminator]{
        \includegraphics[scale=0.35]{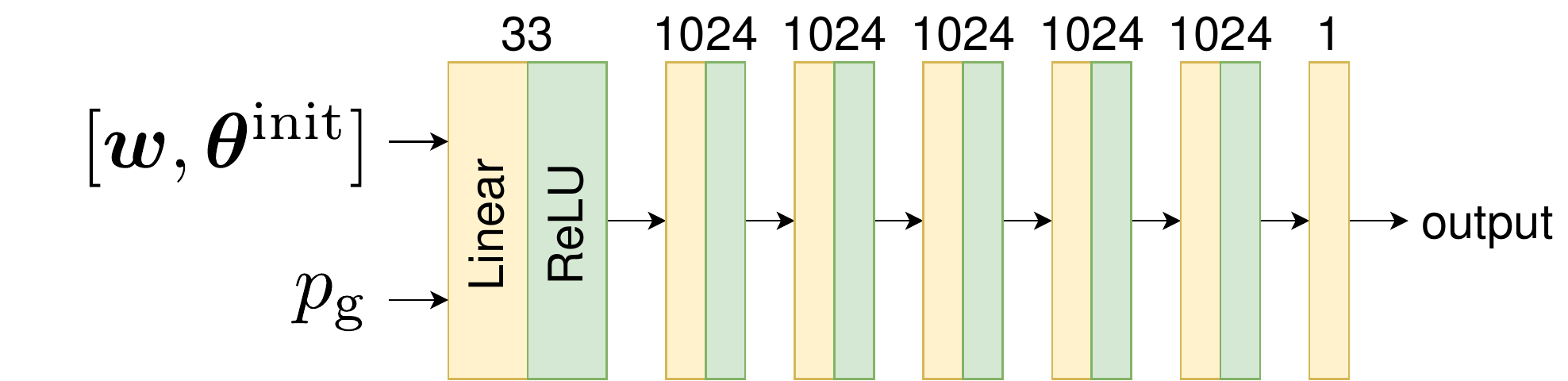}
        \label{fig:discriminator}}
    \caption{Architecture of cWGAN-GP.}
    \label{fig:model architecture}
\end{figure}

We used a cWGAN-GP with the architecture illustrated in Fig.~\ref{fig:model architecture}.
The generator receives a latent variable $\bm{z}$ taken from a uniform distribution and a goal distance $p_\mathrm{g}$.
Then, it outputs the initial pose $\bm{\theta}^\mathrm{init}$ and the weights of the motion primitives $\bm{w}$.
The discriminator receives the initial pose $\bm{\theta}^\mathrm{init}$, the action $\bm{w}$, and the goal distance $p_\mathrm{g}$.
Then, it outputs a scalar to distinguish the dataset and the generated values. 
The hyperparameters for training are detailed in Table~\ref{tbl:hyperparameters of GAN}.

\subsection{Results}
\begin{figure}
    \centering
    \includegraphics[width=7cm]{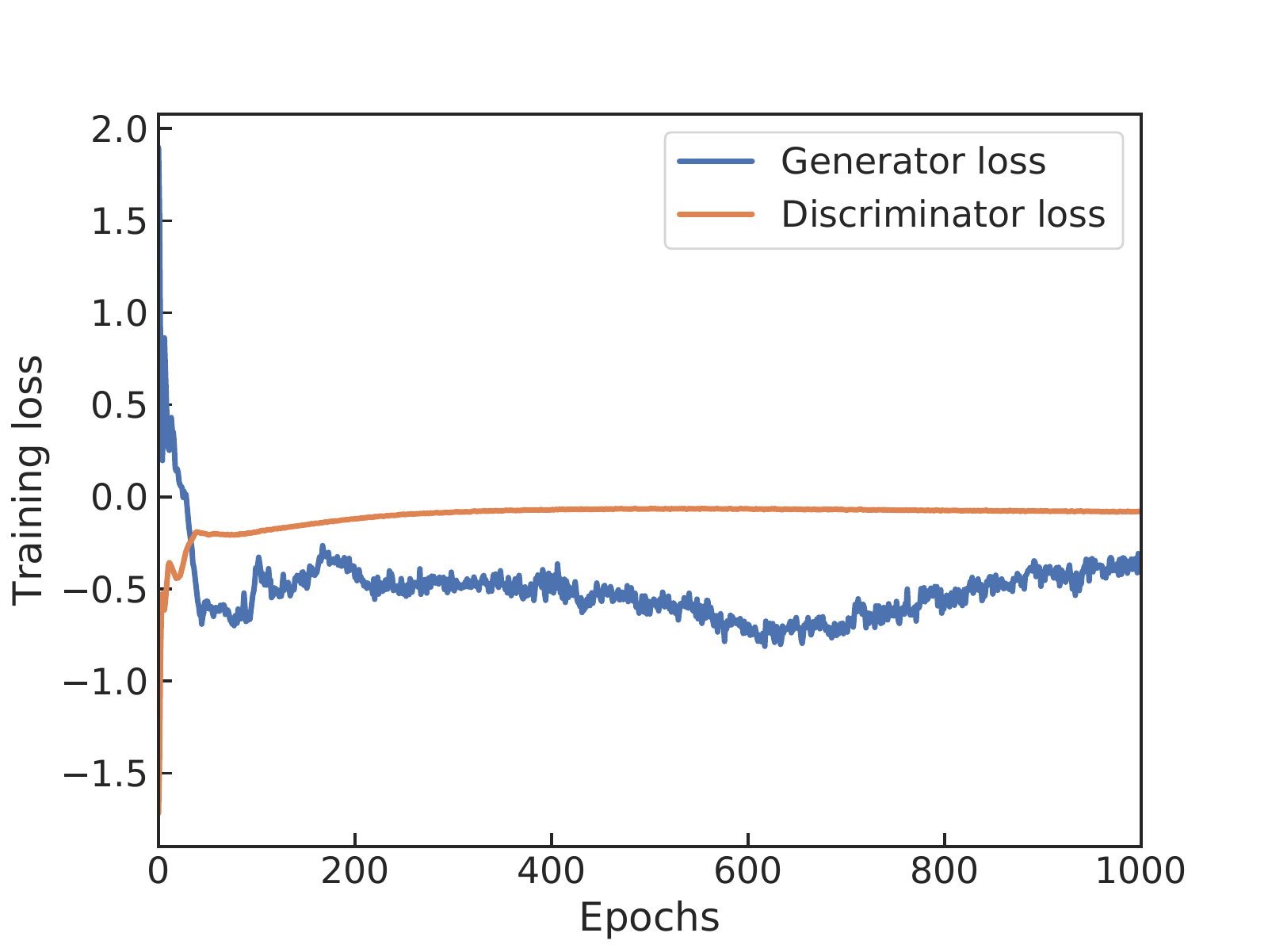}
    \caption{Training losses.}
    \label{fig:training loss}
\end{figure}

\begin{figure}
    \centering
    \subfloat[Success cases]{
        \includegraphics[width=8cm]{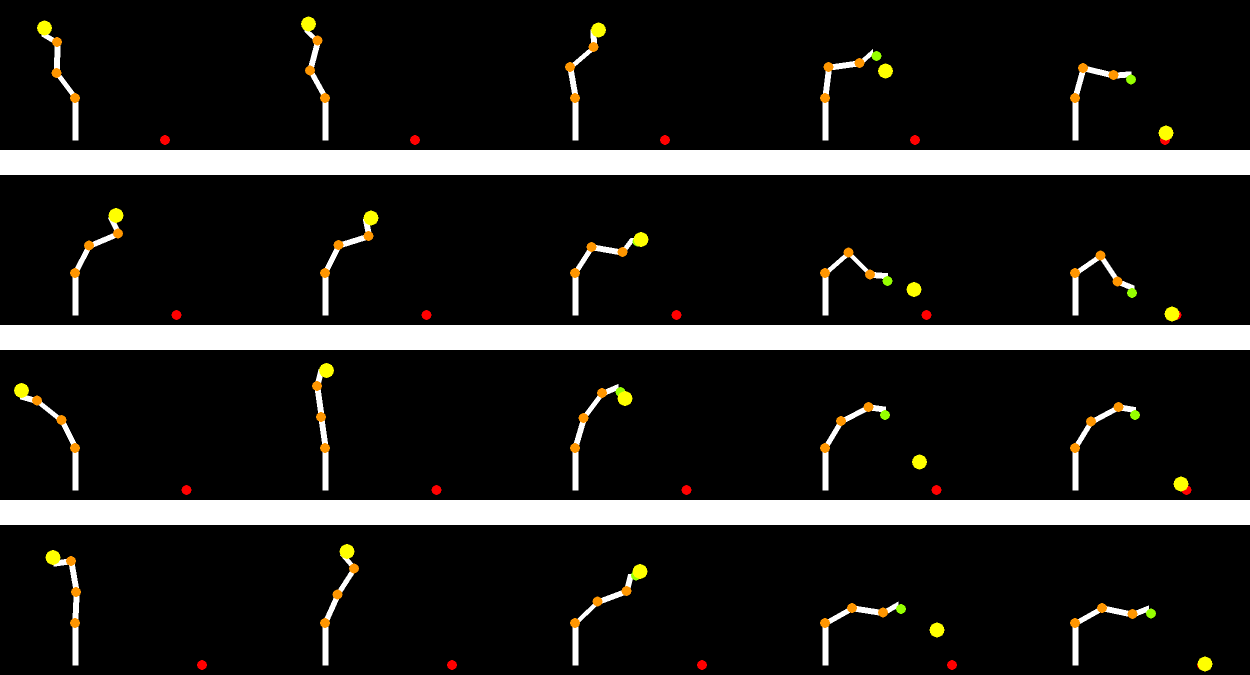}
        \label{fig:succeeded case}}\\
    \subfloat[Failure case]{
        \includegraphics[width=8cm]{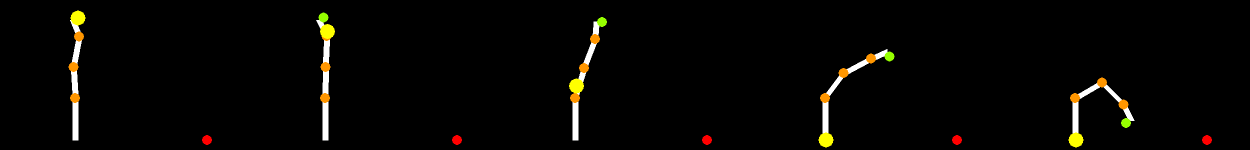}
        \label{fig:failure case}}\\
    \caption{Snapshots of generated motions.
        The trained cWGAN-GP generated various motions.
        In addition, the robot succeeded in throwing near the target positions in most cases.
        In some cases, the robot dropped the object and failed in the task.}
    \label{fig:snapshots of simulation}
\end{figure}

The training losses of the generator and discriminator are shown in Fig.~\ref{fig:training loss}.
We generated throwing motions with the trained cWGAN-GP by sampling the latent variable $\bm{z}$ and the target position $p_\mathrm{g}$ from the uniform distribution.
Snapshots of throwing motions generated by the trained cWGAN-GP are shown in Fig.~\ref{fig:snapshots of simulation}.
Various throwing motions were observed.
In addition, the object was thrown to the target position in most cases.
In some cases, however, the robot dropped the object resulted in large error as shown in Fig.~\ref{fig:snapshots of simulation}\subref{fig:failure case}.
This is considered to be caused by the errors of the acceleration, which result in releasing the object at an undesired timing.

\begin{figure}
    \centering
    \includegraphics[width=7cm]{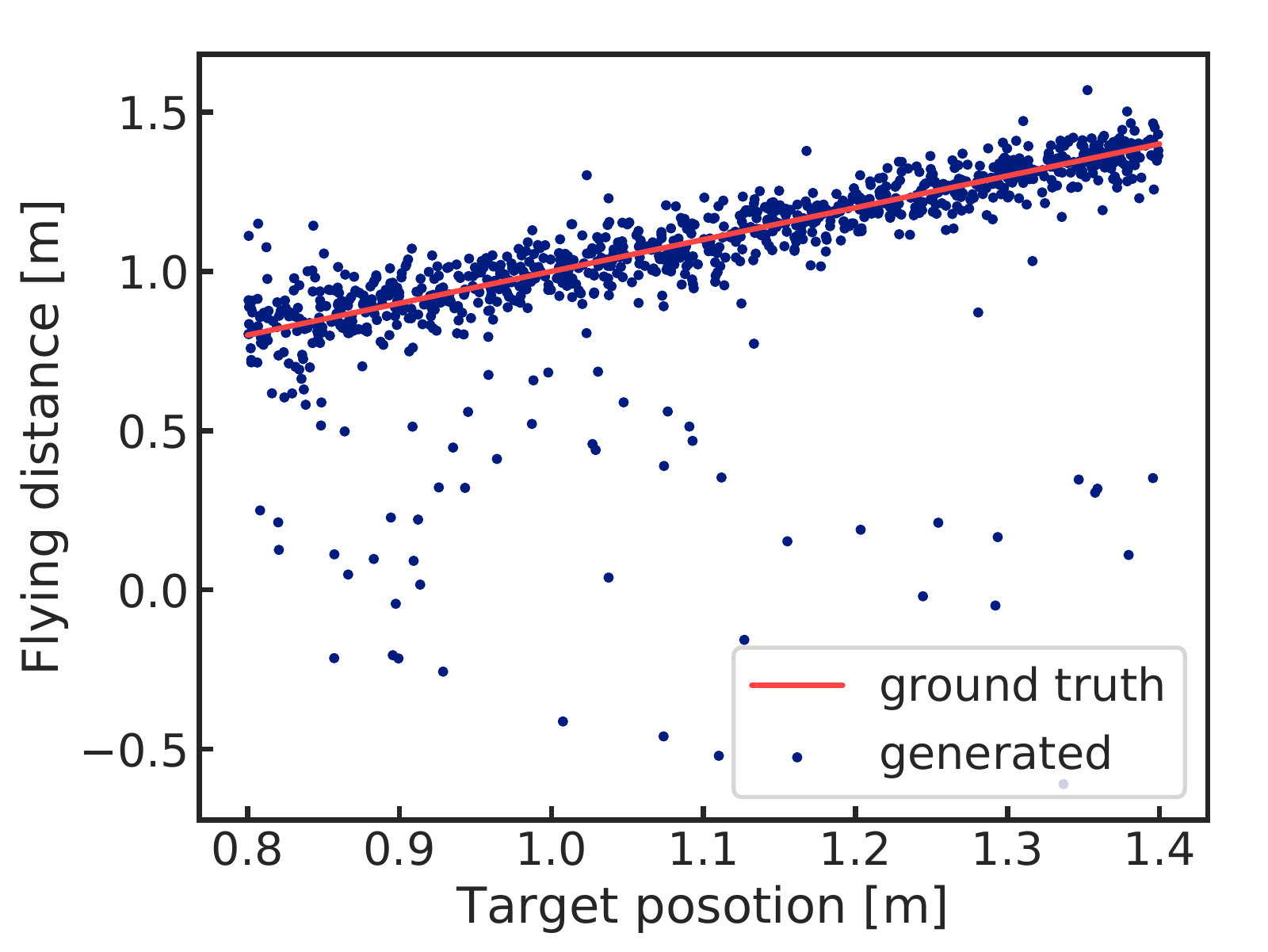}
    \caption{Relationship between the target positions and the flying distance.
        Although small errors remain, the trained cWGAN-GP generated motions mostly land near the target positions.}
    \label{fig:result of accuracy}
\end{figure}

The relationship between the target position and the flying distance by the generated throwing motions is shown in Fig.~\ref{fig:result of accuracy}.
We evaluated 1000 motions with the target positions 0.8--1.4 m.
The generated motions were able to throw the objects near the target positions.
The average error between the target positions and the flying distance was 9 \% (9 cm).

\begin{table}[tb]
    \centering
    \caption{Limitations of the Joints}
    \label{tbl:limitations}
    \begin{tabular}{lccc}
        \hline
        & joint 0 & joint 1 & joint 2\\
        \hline
        Min angle [deg] & $-45$ & $-45$ & $-90$ \\
        Max angle [deg] & $105$ & $105$ & $ 90$ \\
        \hline
        Min velocity [deg/s] & $-120$ & $-180$ & $-180$ \\
        Max velocity [deg/s] & $120$ & $180$ & $180$ \\
        \hline
    \end{tabular}
\end{table}


\begin{table}[tb]
    \centering
    \caption{Hyperparameters for Training}
    \label{tbl:hyperparameters of GAN}
    \begin{tabular}{cc}
        \hline
        Item & Value \\
        \hline
        \# training data &  $282500$\\
        \# epochs &  $1000$\\
        batchsize &  $1024$\\
        $\alpha$ &  $1 \times 10^{-5}$\\
        $\beta_1$ & $0$\\
        $\beta_2$ & $0.5$\\
        $\lambda$ & $10$\\
        \hline
    \end{tabular}
\end{table}


\section{Simulation Results} \label{sec:simulation}
Here the results of the proposed method are described.

\subsection{Setup}
We consider cases that the robot should throw objects within the given workspace.
Such constraints for the situations are implemented as objective functions of CMA-ES.
Here, two examples are verified.

The first objective function is designed as follows:
\begin{align}
    l_1 = \sum_{t=0}^{T-1} \big[& 10 l_{x}(\bm{x}(t)) + l_{\dot{x}} (\dot{\bm{x}}(t))
        + l_\mathrm{flr}(\bm{q}(t)) + l_\mathrm{lim1}(\bm{q}(t)) \big] \label{eqn:task1},
\end{align}
where $\bm{q}(t)$ indicates the positions of joints and the end-effector.
$l_{x}$ and $l_{\dot{x}}$ are the penalty for joint angles and joint angular velocity, respectively.
These are hinge functions whose values increase as the states exceed the limitation.
Here, the limit of the joint angle was the same as Table~\ref{tbl:limitations}, while the limit of the angular velocity was narrowed to $\pm 1.5$ rad, $\pm 1.5$ rad, and $\pm 1.3$ rad for joint 0, 1, and 2, respectively.
That is to avoid a large control deviation when the actual robot reproduces the throwing motions.
$l_\mathrm{flr}$ is also a hinge function whose value increases when the robot came below 0.2 m.
This penalty aims to avoid the robot hitting the floor.
$l_\mathrm{lim1}$ is a limitation of the range of the motions.
It increases when the robot exceeded 0.1 m behind.

The second objective function is designed as follows:
\begin{align}
    l_2 = \sum_{t=0}^{T-1} \big[& 10 l_{x}(\bm{x}(t)) + l_{\dot{x}} (\dot{\bm{x}}(t))
        + l_\mathrm{flr}(\bm{q}(t)) + l_\mathrm{lim2}(\bm{q}(t)) \big] \label{eqn:task2}.
\end{align}
Here, $l_\mathrm{lim2}$ is a limitation of the range of the motions.
The value increases when the robot exceeded 0.5 m in front.

It should be noted that the objective functions do not require any penalty terms related to the target reaching.
This is because the flying distance can be specified as a condition to the trained cWGAN-GP.

The number of samples that CMA-ES samples at each iteration was set to 64.
The initial parameters of the searching distribution were set to a mean of 0.0 and the standard deviation of 0.4.
The search was continued until the value of the objective function is reached zero.

\subsection{Results}
\begin{figure}
    \centering
    \subfloat[the throwing motion obtained by $l_1$.]{
        \includegraphics[width=8cm]{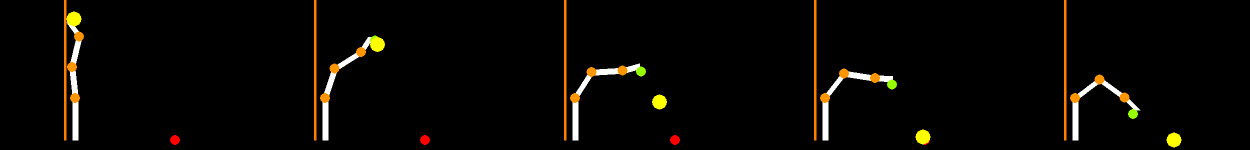}
        \label{fig:snapshot task 1}
    }\\
    \subfloat[the throwing motion obtained by $l_2$.]{
        \includegraphics[width=8cm]{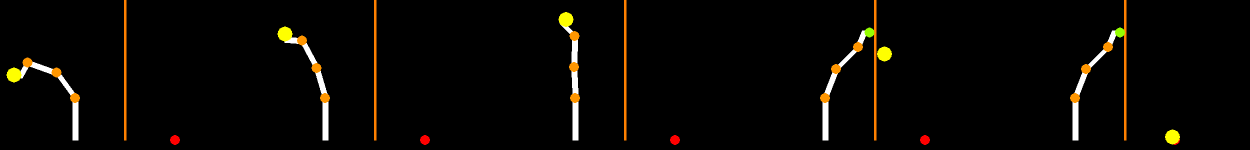}
        \label{fig:snapshot task 2}
    }\\
    \caption{Snapshots of the throwing motion obtained by searching the latent space.
        The orange line indicates the boundary of the motion range.}
    \label{fig:results of latent search}
\end{figure}

The obtained motion by $l_1$ is shown in Fig.~\ref{fig:results of latent search}\subref{fig:snapshot task 1}.
Robots did not hit to the wall during the motion.

The obtained motion by $l_2$ is shown in Fig.~\ref{fig:results of latent search}\subref{fig:snapshot task 2}.
A different motion was obtained compared with Fig.~\ref{fig:snapshot task 1}.
Robots did not hit to the obstacle during the motion.

In both cases, the values of the objective functions were zero.

\subsection{Comparison with direct search in the action space}
To verify the effectiveness of searching the latent space, we compare it with searching the action space.

Here, the action space consists of $\bm{x}(0)$ and $\bm{w}$, which are the same as the output of the cWGAN-GP.
Its dimension is 33: the initial pose $\bm{x}(0)$ and the weights of the motions $\bm{w}$.

For comparison, we used the following objective function:
\begin{align}
    l = \sum_{t=0}^{T-1} \big[& 10 l_{x}(\bm{x}(t)) + l_{\dot{x}} (\dot{\bm{x}}(t)) + l_\mathrm{flr}(\bm{q}(t)) + l_\mathrm{lim1}(\bm{q}(t)) \nonumber\\
        &+ \max(\|p_x(T-1) - p_\mathrm{g}\|-0.1, 0) \big].
\end{align}
Here, $p_x(T-1)$ is the landing position of the object.
$p_\mathrm{g}$ is set to 1.0 m in this case.
This is similar to $l_1$ in (\ref{eqn:task1}), while a term for flying distance is added.

\begin{figure}
    \centering
    \subfloat[searching the action space (landing position: 80 cm)]{
        \includegraphics[width=8cm]{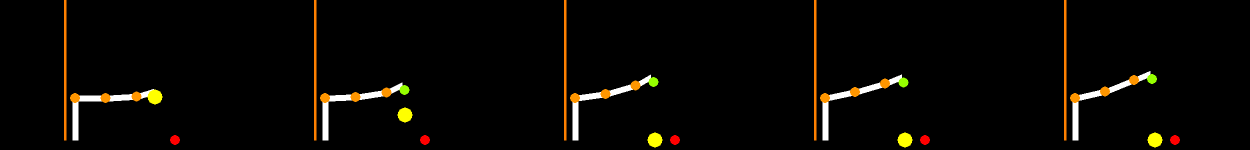}
        \label{fig:snapshot action space search}
    }\\
    \subfloat[searching the latent space (landing position: 94 cm)]{
        \includegraphics[width=8cm]{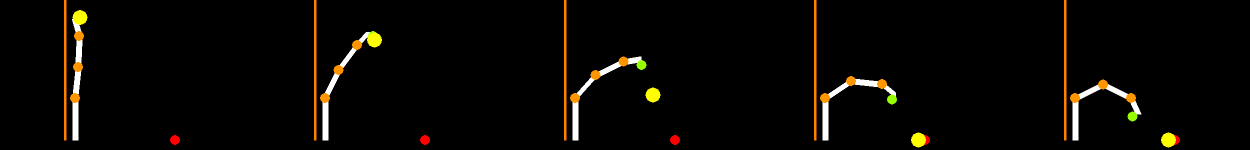}
        \label{fig:snapshot latent space search}
    }\\
    \caption{Snapshots of the comparison of the searching space.
        }
    \label{fig:comparison in searching space}
\end{figure}

At first, we searched in the action space directly.
As a result, we could not reach the value for the objective function of zero.
The value converged to 0.28350 in the 64th update of CMA-ES in 796 s.
The landing position of the object was 80 cm, while the target position was 100 cm.
Therefore, the error was 20 \%.
Snapshots of the obtained motion are shown in Fig.~\ref{fig:comparison in searching space}\subref{fig:snapshot action space search}.
The motion seems to just extend the arm to the limit and drop the object.
Such motion cannot carry the object beyond the reachable space.
We conducted the same evaluation five times and each resulted in the same kind of failure.

On the other hand, the proposed method found the solution that makes the value of the objective function zero in the 9th update in 295 s.
The landing distance of the object was 94 cm, that is, the error was 6 \%.
Therefore, the proposed method resulted in over three times higher accuracy in about 40 \% of the calculation time from searching the action space directly.
The snapshots are shown in Fig.~\ref{fig:comparison in searching space}\subref{fig:snapshot latent space search}.
We conducted the same evaluation five times, with successful throwing motions in all trials.

\section{Real Robot Experiments} \label{sec:experiment}
To verify that the proposed method also works in actual robots, we conducted the real robot experiments.

\subsection{Setup}
\begin{figure}
    \centering
    \subfloat[Manipulator]{
        \includegraphics[width=3cm]{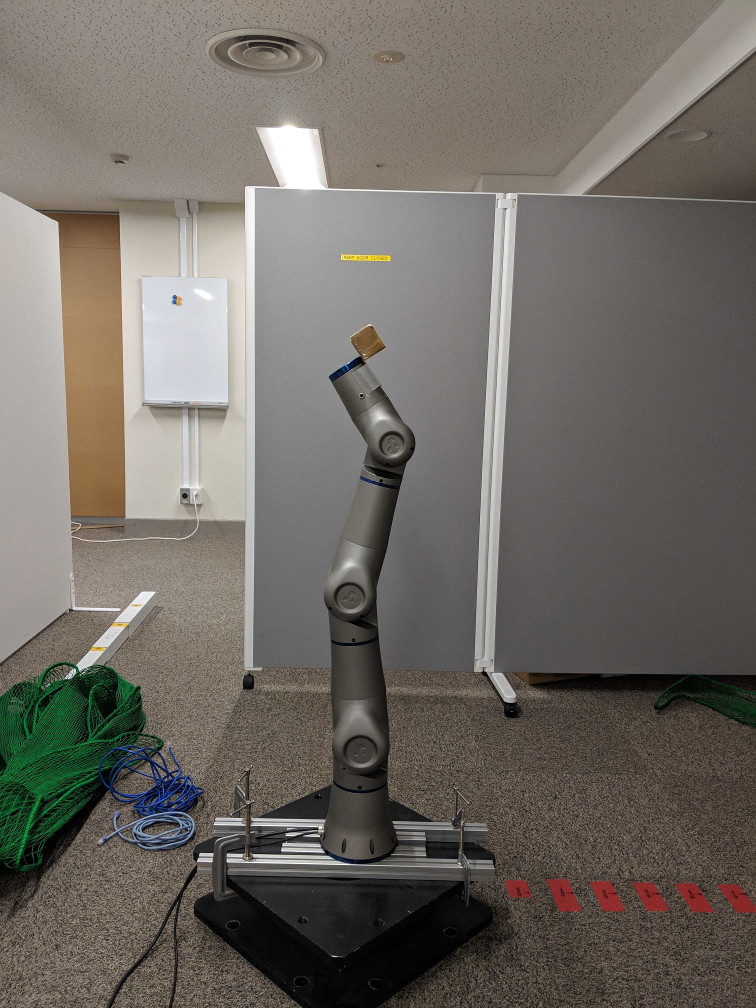}
        \label{fig:toroboarm}
    }
    \subfloat[End-effector]{
        \includegraphics[width=3cm]{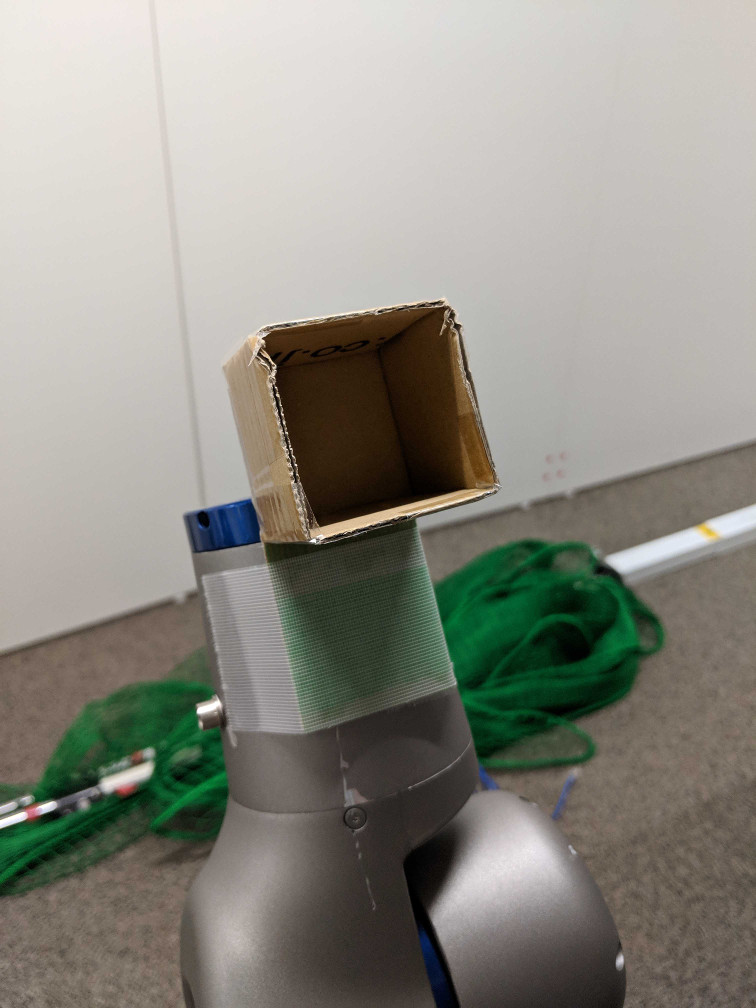}
        \label{fig:end-effector}
    }
    \caption{Experimental setup.}
    \label{fig:experimental setup}
\end{figure}

We used a seven degrees of freedom robot arm, ToroboArm, supplied by Tokyo Robotics.
Its overview is shown in Fig.~\ref{fig:experimental setup}\subref{fig:toroboarm}.
Although it has seven joints, we used only three joints for throwing motions.
Each joint is controlled based on control commands of the angle, angular velocity, and angular acceleration.

The robot was equipped with an end-effector as shown in Fig.~\ref{fig:experimental setup}\subref{fig:end-effector}.

\subsection{Results}
\begin{figure*}
    \centering
    \subfloat[Trajectory 2 (actual flying distance: 101.8 cm)]{
        \includegraphics[width=16cm]{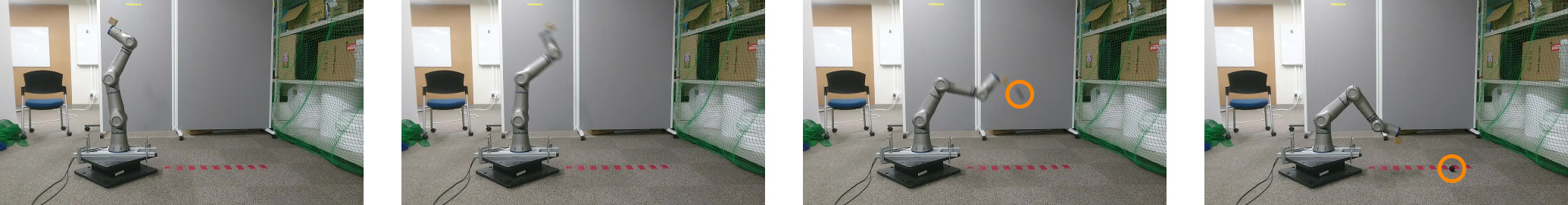}
    }\\
    \subfloat[Trajectory 6 (actual flying distance: 117.4 cm)]{
        \includegraphics[width=16cm]{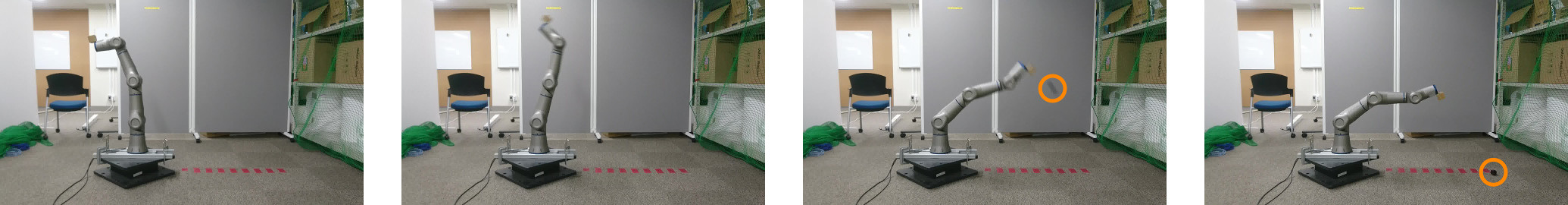}
    }\\
    \caption{Snapshots of throwing motions by the actual robot.}
    \label{fig:snapshots of experiment 1}
\end{figure*}

At first, we evaluated throwing motions generated by the trained cWGAN-GP by the actual robot.
The results are described in Table~\ref{tab:results of experiment 1}.
The snapshots are shown in Fig.~\ref{fig:snapshots of experiment 1}.
In most cases, the landing distance was almost the same as the simulation.
The largest error was about 7 \%.
The cause for the errors is believed to be modeling errors of the end-effector and control deviations from the generated motions.

\begin{figure*}
    \centering
    \subfloat[Trajectory 7, obtained by $l_1$ (actual flying distance: 115.0 cm)]{
        \includegraphics[width=16cm]{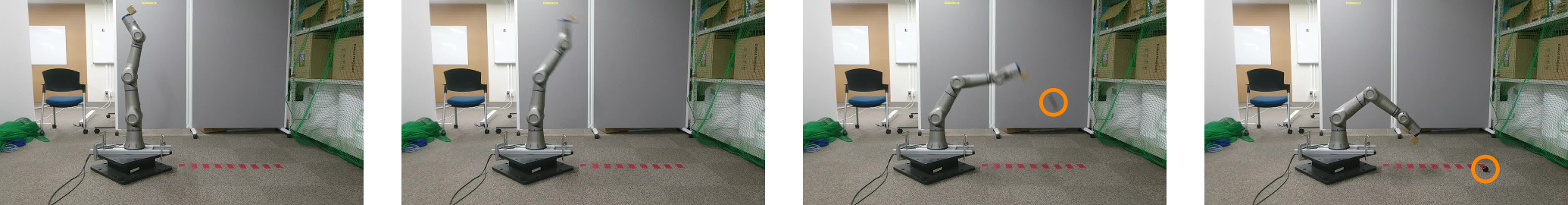}
    }\\
    \subfloat[Trajectory 8, obtained by $l_1$ (actual flying distance: 104.1 cm)]{
        \includegraphics[width=16cm]{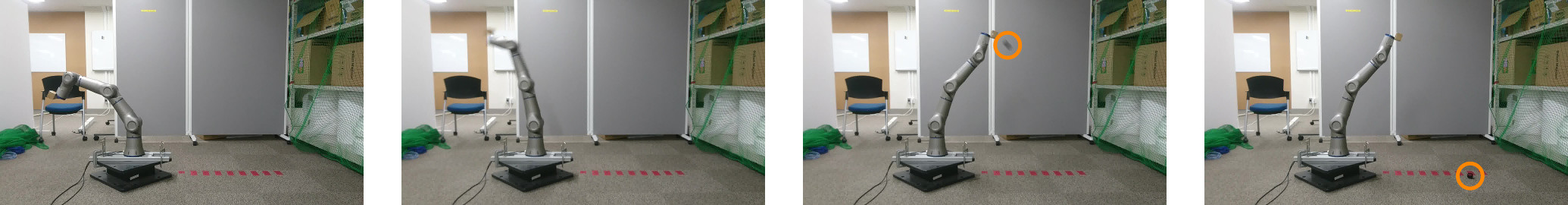}
    }
    \caption{Snapshots that the actual robot executes the throwing motion obtained by searching the latent space.}
    \label{fig:snapshots of experiment 2}
\end{figure*}

Next, we evaluated motions subject to movement restrictions obtained in Section~\ref{sec:simulation}.
The results are described in Table~\ref{tab:results of experiment 2}.
Snapshots are shown in Fig.~\ref{fig:snapshots of experiment 2}.
The deviations to the simulation were larger than the above results.
We believe that the modeling errors of the contact model of the end-effector appeared due to the pose for object avoidance.

\begin{table}[tb]
    \centering
    \caption{Experimental results of throwing motions}
    \label{tab:results of experiment 1}
    \begin{tabular}{cccc}
        \hline
        No. & Target & Simulation & Actual robot \\
        \hline
        1 & $100.0$ cm & $107.3$ cm & $108.7 \pm 0.5$ cm \\
        2 & $100.0$ cm &  $98.8$ cm & $101.8 \pm 0.6$ cm \\
        3 & $100.0$ cm & $100.6$ cm & $107.3 \pm 0.7$ cm \\
        \hline
        4 & $120.0$ cm & $117.6$ cm & $115.1 \pm 0.5$ cm \\
        5 & $120.0$ cm & $112.8$ cm & $115.5 \pm 0.6$ cm \\
        6 & $120.0$ cm & $117.5$ cm & $117.4 \pm 1.7$ cm \\
        \hline
    \end{tabular}
\end{table}

\begin{table}[tb]
    \centering
    \caption{Experimental results of throwing motions obtained by searching the latent space}
    \label{tab:results of experiment 2}
    \begin{tabular}{cccc}
        \hline
        No. & Target & Simulation & Actual robot \\
        \hline
        7 & $100.0$ cm &  $99.1$ cm & $113.8 \pm 0.9$ cm \\
        8 & $100.0$ cm &  $99.0$ cm & $104.5 \pm 1.3$ cm \\
        \hline
    \end{tabular}
\end{table}

\section{Conclusion and Future Work} \label{sec:conclusion}
For robots to work in our daily lives, they will need to adjust their motions depending on surrounding objects even when performing the same task.
In this paper, we proposed a method based on cGANs to tackle this issue.
By searching latent spaces of cGANs that learned various motions, appropriate motions for novel situations can be obtained.
We use robotic throwing as an example.
We showed that the trained cWGAN-GP can generate various throwing motions.
In addition, we verified that the proposed method can find appropriate throwing motions to different situations by simulation and real-robot experiments.
The appropriate throwing motions could be found without considering the flying distance with objective functions thanks to specifying the condition to the cWGAN-GP.
We also observed that the proposed method could avoid poor local optima (i.e., motions not satisfying the objective) by searching the latent space which represents only valid motions.
As the results, the proposed method resulted in higher accuracy with less calculation time than searching the action space directly.

In this paper, we used a two-dimensional simulator to obtain a large amount of training data.
To apply the proposed method to other tasks, there are some future works remained.
In tasks which are difficult to simulate such as picking and walking, we should use actual data.
Also, if three-dimensional motion planning is necessary, the motions will become higher degrees of freedom, which cause sampling efficiency lower.
Next steps would be to look for methods to reduce the amount of training data needed to use the data obtained with the actual environment.

\section*{ACKNOWLEDGMENT}
The authors would like to thank Crissman Loomis and Kohei Hayashi for useful discussions and advice.

\bibliographystyle{IEEEtran}
\bibliography{IEEEabrv,references}


\end{document}